\documentclass{article}

\usepackage{amsmath,amssymb}
\usepackage{booktabs,graphicx}
\usepackage[final]{corl_2026}

\title{NF-VLA: A Normalizing-Flow Baseline for Vision-Language-Action Learning}

\author{Ruichen Deng \\
Department of Computer and Information Science\\
University of Pennsylvania \\
\texttt{ruichend@seas.upenn.edu}
\And
Qinyu Zhao\\
School of Computing\\
Australian National University\\
\texttt{qinyu.zhao@anu.edu.au}
\And
Yong-Hyun Park\\
Department of Computer and Information Science\\
University of Pennsylvania\\
\texttt{park19@seas.upenn.edu}
\And
Mutian Tong\\
Department of Computer and Information Science\\
University of Pennsylvania\\
\texttt{mutiant@seas.upenn.edu}
\And
Liang Zheng\\
School of Computing\\
Australian National University\\
\texttt{liang.zheng@anu.edu.au}
\And
Jiatao Gu\\
Department of Computer and Information Science\\
University of Pennsylvania\\
\texttt{jgu32@seas.upenn.edu}
}

\begin{document}
\maketitle

\begin{abstract}
Autoregressive normalizing flows (NFs) provide a tractable density for learning and sampling continuous robot actions. We show that a single maximum-likelihood objective can train vision--language--action (VLA) policies competitive with leading methods on manipulation benchmarks. We present NF-VLA, which combines conditional autoregressive normalizing flows with score-based action refinement. Its deep--shallow architecture concentrates visual, language, and proprioceptive prefix conditioning in a high-capacity deep block, while unconditioned shallow blocks provide a shared invertible action transformation. The vision--language backbone and flow are jointly trained by exact maximum likelihood on Gaussian-smoothed action chunks. The same learned density supplies a single score-based correction at inference, refining generated actions to account for training-time smoothing without a separately trained denoiser. Trained on LIBERO from a pretrained vision--language backbone without robot pretraining, NF-VLA achieves 98.35\% success across the four standard suites and 72.6\% on LIBERO-Plus using the same checkpoint without further fine-tuning. Ablations show that deep-only prefix conditioning and score-based refinement both improve closed-loop success. Together, these results establish carefully structured autoregressive flows as a competitive approach to VLA control, with one learned density supporting policy training, action generation, and refinement.
\end{abstract}

\section{Introduction}
\label{sec:introduction}

Vision-language-action (VLA) models connect pretrained visual and linguistic representations to robot control, allowing policies to generate actions from images and language instructions~\cite{brohan2023rt2,kim2024openvla}.
A central design choice is the action model: it determines how observations condition the action distribution and how that distribution produces executable commands.
Existing approaches include discrete action prediction, continuous regression, diffusion, and flow matching~\cite{kim2024openvla,kim2025oft,chi2023diffusionpolicy,black2024pi0}.

Normalizing flows (NFs) offer a formulation in which an invertible transformation defines both a continuous density and a sampling procedure.
Their tractable change-of-variables likelihood allows direct maximum-likelihood training~\cite{dinh2016density,papamakarios2017maf}.
TARFlow~\cite{zhai2025tarflow} and the STARFlow family~\cite{gu2025starflow,gu2025starflowv,shen2026starflow2} demonstrate the potential of Transformer autoregressive flows for visual and multimodal generation, with STARFlow introducing a deep--shallow architecture for scaling latent image generation.
These developments motivate a study of conditioning and sampling in autoregressive-flow VLA policies: where and how to inject multimodal context, and whether the learned density's score can improve actions after sampling.

Recent work establishes normalizing flows as capable robot policies. NinA~\cite{tarasov2025nina} introduces coupling-based action decoders for VLAs, emphasizing efficient generation, while RoMAN-Flow~\cite{romanflow} develops autoregressive-flow policies for offline reinforcement learning and one-step policy distillation. Building on this progress, we focus on the manipulation performance obtainable from likelihood-based imitation learning itself. The central design question is how to organize multimodal conditioning and sample refinement within a single continuous-density model.

We present \textbf{NF-VLA}, a VLA policy that combines conditional autoregressive flows with score-based action refinement. Building on the deep--shallow architecture of STARFlow~\cite{gu2025starflow}, we concentrate visual, language, and proprioceptive conditioning in a high-capacity deep block, while shallow blocks provide a shared invertible transformation between action coordinates and intermediate latents.
The vision--language backbone and action model are trained jointly with a single exact maximum-likelihood objective on Gaussian-smoothed actions.
At inference, the same learned density supports flow inversion and a single Tweedie denoising correction computed from its own score~\cite{efron2011tweedie,zhai2025tarflow}.
Learning, sampling, and refinement therefore share one density model, without a separately trained denoiser or an additional training objective.

Starting from a pretrained vision--language backbone, NF-VLA is trained jointly on the full LIBERO dataset without robot pretraining or reinforcement-learning post-training. It achieves \textbf{98.35\%} success across the four standard LIBERO suites~\cite{liu2023libero}, including \textbf{97.8\%} on LIBERO-Long. The same checkpoint achieves \textbf{72.6\%} on LIBERO-Plus~\cite{fei2025liberoplus} without further fine-tuning, demonstrating strong aggregate performance under benchmark perturbations. These results place NF-VLA competitively alongside leading VLA policies in the benchmark comparisons. Ablations support the two design choices: concentrating prefix conditioning in the deep block improves average LIBERO success by 1.20 percentage points over all-block prefix conditioning, and a single score-based refinement step improves success by 1.90 percentage points using the same trained checkpoint.

Our contributions are threefold:
\begin{itemize}
    \item We present \textbf{NF-VLA}, an autoregressive-flow VLA baseline trained end to end with a single likelihood objective and using the learned density for both sampling and score-based action correction.
    \item We evaluate conditioning placement and mechanism, finding that deep-only prefix conditioning achieves higher average success than all-block conditioning in our tested configurations. We additionally quantify the control benefit of self-score correction and report the inference costs of the evaluated designs.
    \item We evaluate the resulting policy on LIBERO and LIBERO-Plus, reporting strong standard-task performance alongside a perturbation-dependent robustness profile.
\end{itemize}

\section{Related Work}
\label{sec:related_work}

\subsection{Vision-Language-Action Models}
\label{sec:related_vla}

Vision-language-action (VLA) models adapt pretrained vision--language representations to language-conditioned robot control.
RT-2~\cite{brohan2023rt2} represents robot actions as text tokens, allowing vision--language models to learn control alongside web-based vision--language tasks.
OpenVLA~\cite{kim2024openvla} develops an open-source VLA with discretized action prediction and demonstrates adaptation to new manipulation tasks.
Subsequent work explores alternatives to token-by-token action decoding: OpenVLA-OFT~\cite{kim2025oft} combines parallel decoding and action chunking with continuous-action regression, showing that action representation and training objectives are consequential design choices.

Generative action models provide another approach to continuous control.
Diffusion Policy~\cite{chi2023diffusionpolicy} models action sequences through conditional denoising diffusion and executes them with receding-horizon control.
Within VLAs, $\pi_0$~\cite{black2024pi0} couples a pretrained vision--language model with a flow-matching action expert, while SmolVLA~\cite{shukor2025smolvla} develops a compact architecture with a flow-matching decoder.

\subsection{Normalizing Flows}
\label{sec:related_flows}

NFs model continuous densities through invertible transformations with tractable Jacobian determinants.
Real NVP~\cite{dinh2016density} uses affine coupling layers, while inverse autoregressive flow~\cite{kingma2016improved} and masked autoregressive flow~\cite{papamakarios2017maf} exploit autoregressive structure to obtain triangular Jacobians.
More recently, TARFlow~\cite{zhai2025tarflow} combines affine autoregressive transformations with Transformers for image generation, and the STARFlow family~\cite{gu2025starflow,gu2025starflowv,shen2026starflow2} develops latent-space flows with deep--shallow architectures for visual and multimodal generation.
Flow matching~\cite{lipman2022flowmatching} is also connected to NFs: it trains continuous-time flows by regressing vector fields along probability paths.

Recently in robotics, Ghugare and Eysenbach~\cite{nfcontrol} investigate their use in behavior cloning and reinforcement learning. Within VLAs, NinA~\cite{tarasov2025nina} replaces FLOWER's action decoder with Real NVP-style coupling flows~\cite{dinh2016density}, evaluating MLP and Transformer conditioners and emphasizing inference efficiency. RoMAN-Flow~\cite{romanflow} uses autoregressive flows with context supplied to every flow block, followed by advantage-weighted offline policy optimization and one-step distillation. SERNF~\cite{sernf} combines normalizing-flow policies with action-chunked critics for off-policy fine-tuning on real robots. Our focus is the performance of the likelihood-trained VLA policy itself: we concentrate direct conditioning in the deep flow block and reuse its action density for inference-time refinement, retaining a single maximum-likelihood training objective.

\section{Preliminaries}
\label{sec:preliminaries}

\subsection{Normalizing Flows}
\label{sec:normalizing_flows}

An NF represents a continuous probability distribution through an invertible transformation with a tractable Jacobian determinant~\cite{dinh2016density,papamakarios2017maf,zhai2025tarflow}.
For a variable $\mathbf{x}\in\mathbb{R}^{n}$ and conditioning information $\mathbf{c}$, let $f_{\theta}(\cdot;\mathbf{c})$ map data to a latent variable $\mathbf{z}$ with standard Gaussian density $p_0$.
The change-of-variables formula gives
\begin{equation}
\begin{aligned}
\mathbf{z} &= f_{\theta}(\mathbf{x};\mathbf{c}),\\
\log q_{\theta}(\mathbf{x}\mid\mathbf{c})
&= \log p_0(\mathbf{z})
+ \log\left|\det\frac{\partial f_{\theta}(\mathbf{x};\mathbf{c})}{\partial\mathbf{x}}\right|.
\end{aligned}
\label{eq:change_of_variables}
\end{equation}
The Jacobian is taken with respect to $\mathbf{x}$ while holding $\mathbf{c}$ fixed; the conditioning encoder need not be invertible.

For a composition of $K$ invertible blocks, $f_{\theta}=f_K\circ\cdots\circ f_1$, define $\mathbf{x}^{(0)}=\mathbf{x}$ and $\mathbf{x}^{(k)}=f_k(\mathbf{x}^{(k-1)};\mathbf{c})$.
The log-Jacobian determinants add across blocks, yielding the negative log-likelihood
\begin{equation}
\begin{aligned}
-\log q_{\theta}(\mathbf{x}\mid\mathbf{c})
={}&\frac{1}{2}\|\mathbf{z}\|_2^2 -\sum_{k=1}^{K}\log\left|\det\frac{\partial f_k(\mathbf{x}^{(k-1)};\mathbf{c})}{\partial\mathbf{x}^{(k-1)}}\right| + \frac{n}{2}\log(2\pi).
\end{aligned}
\label{eq:flow_nll}
\end{equation}
Maximum-likelihood training minimizes its expectation over the training distribution.
Generation reverses the transformation: sample $\mathbf{z}\sim\mathcal{N}(\mathbf{0},\mathbf{I}_n)$ and compute $\mathbf{x}=f_{\theta}^{-1}(\mathbf{z};\mathbf{c})$.

\subsection{Autoregressive Flows}
\label{sec:autoregressive_flows}

Autoregressive flows make the Jacobian determinant tractable by restricting each coordinate's transformation parameters to depend only on preceding coordinates~\cite{kingma2016improved,papamakarios2017maf}.
For an ordered input $\mathbf{x}=(x_1,\ldots,x_n)$, let $b_i=b_{\theta,i}(\mathbf{x}_{<i};\mathbf{c})$ and $s_i=s_{\theta,i}(\mathbf{x}_{<i};\mathbf{c})$ denote the shift and scale.
An affine autoregressive block and its inverse take the form
\begin{equation}
z_i=\frac{x_i-b_i}{s_i},
\qquad
x_i=b_i+s_i z_i,
\qquad i=1,\ldots,n.
\label{eq:autoregressive_affine}
\end{equation}
The Jacobian is triangular with diagonal entries $1/s_i$, giving
\begin{equation}
\log\left|\det\frac{\partial\mathbf{z}}{\partial\mathbf{x}}\right|
=-\sum_{i=1}^{n}\log s_i.
\label{eq:autoregressive_logdet}
\end{equation}
A causally masked network can compute all coefficients in parallel for an observed input, whereas inversion recovers coordinates sequentially from the generated prefix.
For vector-valued tokens, the affine transformation acts elementwise, with each token's coefficients conditioned on preceding tokens.

Transformer autoregressive flow (TARFlow)~\cite{zhai2025tarflow} parameterizes these coefficients with causal Transformers and stacks blocks with alternating autoregressive orderings.
STARFlow~\cite{gu2025starflow} builds on this formulation with a deep--shallow architecture: in the generative direction, a deep block maps Gaussian noise to intermediate latents, which subsequent shallow blocks transform into model outputs.
Conditioning can be supplied as a prefix to the deep block while the shallow blocks remain unconditioned.
We adopt STARFlow as NF-VLA's autoregressive flow backbone and adapt this architecture to observation-conditioned continuous action chunks.

\section{NF-VLA}
\label{sec:method}

\begin{figure}[t]
\centering
\includegraphics[width=\linewidth]{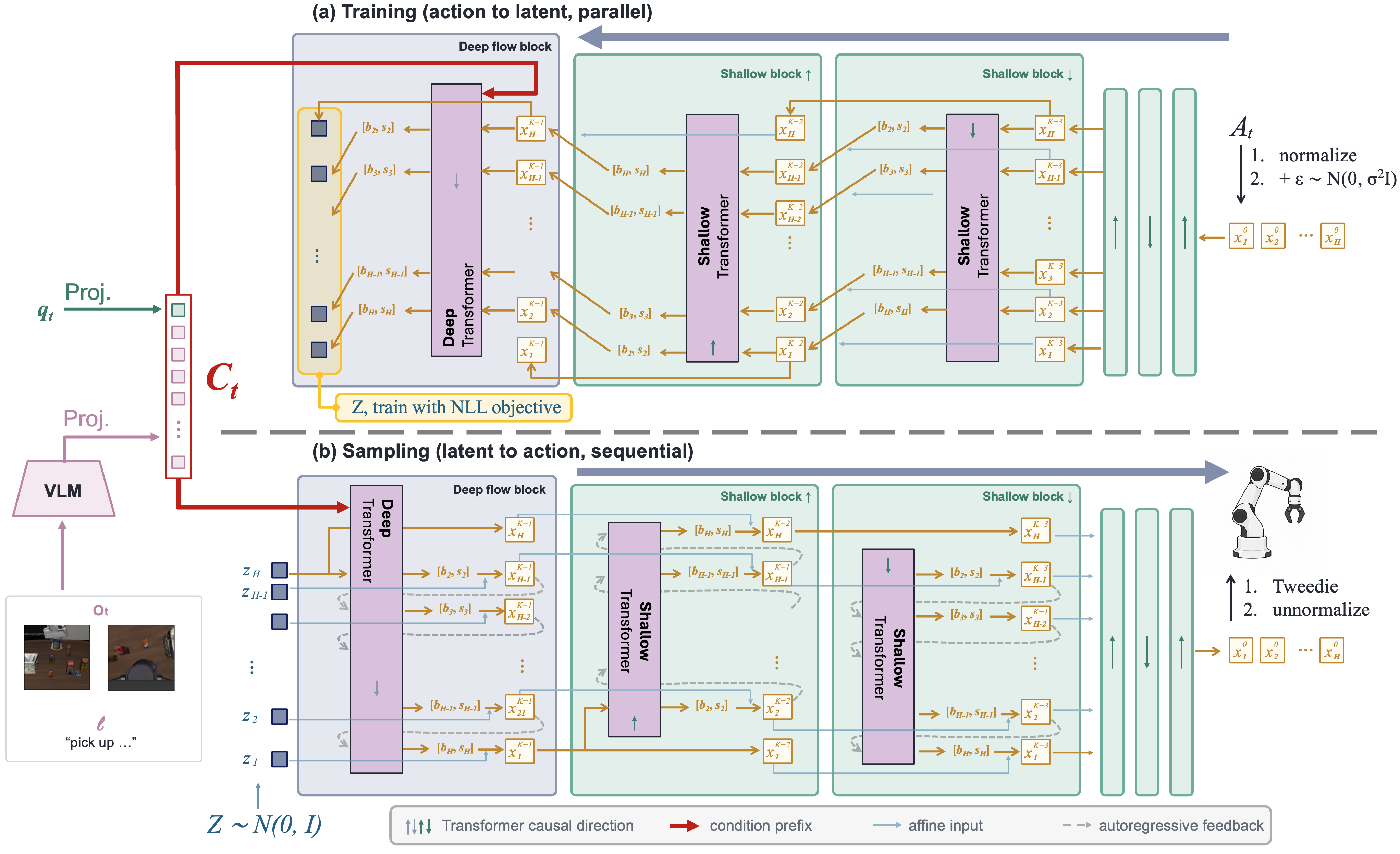}
\caption{\textbf{NF-VLA architecture for training and sampling.} Vision--language features and proprioception are projected into a shared prefix $\mathbf{C}_t$ that conditions only the deep flow block. \textbf{(a)} Training maps normalized, Gaussian-perturbed action chunks through the shallow blocks and the deep block to latents $\mathbf{Z}$, evaluating affine coefficients in parallel and optimizing the exact negative log-likelihood. \textbf{(b)} Sampling reverses the blocks, recovering tokens autoregressively from Gaussian noise before Tweedie denoising and action unnormalization.}
\label{fig:architecture}
\end{figure}

NF-VLA combines a pretrained vision--language backbone with a conditional NF over continuous action chunks  (Figure~\ref{fig:architecture}).
Visual, language, and proprioceptive features condition a deep autoregressive block, while shallow flow blocks transform between action latents and normalized actions.
We train the model end to end by exact maximum likelihood on Gaussian-smoothed actions and generate actions by flow inversion followed by a single score-based denoising step.

\subsection{Action representation and multi-modal conditioning}
\label{sec:conditioning}

At time $t$, the policy receives observations $\mathbf{o}_t=(\mathbf{I}^{1}_t,\mathbf{I}^{2}_t, \cdots, \mathbf{I}^{n}_t)$, a language instruction $\ell$, and proprioception $\mathbf{q}_t$ and generates a normalized action chunk with horizon $H$ and action dimension $d_a$.
The instruction and observations are encoded with a trainable pretrained VLM~\cite{marafioti2025smolvlm}, producing embeddings
\begin{equation}
\mathbf{H}_t = \text{VLM}_\phi(\mathbf{o}_t, \ell) \in \mathbb{R}^{N_{v}\times d_{v}}
\end{equation}
Two learned linear projections map normalized proprioception $\bar{\mathbf{q}}_t$ and the vision--language features to a common width $d$, forming the multi-modal condition:
\begin{equation}
\mathbf{C}_t=
\operatorname{Concat}_{\mathrm{tokens}}\!\left(
\operatorname{Proj}_q(\bar{\mathbf{q}}_t),
\operatorname{Proj}_v(\mathbf{H}_t)
\right)
\in\mathbb{R}^{L_c\times d},
\label{eq:context}
\end{equation}
where $L_c = N_v + 1$.

\subsection{Context-conditioned autoregressive flow}
\label{sec:flow_architecture}
NF-VLA adopts the TARFlow parameterization~\cite{zhai2025tarflow} with the deep--shallow architecture of STARFlow~\cite{gu2025starflow}. The head comprises $K$ affine flow blocks with alternating autoregressive ordering: in the forward pass, the normalized action first goes through unconditioned shallow blocks, then maps to the latent through a deep block conditioned with $\mathbf{C}_t$. The reverse pass therefore begins with the conditioned deep block and proceeds through the unconditioned shallow blocks. The architectural dimensions are specified in Section~\ref{sec:implementation}.

The deep block provides a high-capacity component for capturing task-relevant conditional dependencies in a shared action-latent space.
Keeping the shallow stack free of direct context inputs concentrates conditioning in this component and restricts additional context-dependent reparameterizations across flow blocks.
This structural constraint may simplify learning by reducing unnecessary inter-block co-adaptation.

\paragraph{Attention mask.}
In the deep block, projected action tokens follow the context tokens $\mathbf{C}_t$ in a shared Transformer residual stream.
Context tokens attend bidirectionally within the context prefix but not to action tokens.
Each action token attends to all context tokens, preceding action tokens, and itself, with future action positions masked.
We shift the raw affine parameters right by one action position and prepend zeros before applying Eqs.~\ref{eq:tanhclip} and Eqs.~\ref{eq:softplus}.
Thus, the coefficients for token $i\geq2$ come from the Transformer output at position $i-1$, which sees only preceding tokens.
Shallow blocks use the same inclusive causal action mask and output shift, without context tokens.

\paragraph{Invertible action transformations.}
Let $\mathbf{X}^{(0)}$ denote the flow input.
For block $k\in\{1,\ldots,K\}$, the permutation $P_k$ preserves token order for odd $k$ and reverses it for even $k$.
The block forms $\mathbf{U}^{(k)}=P_k\mathbf{X}^{(k-1)}$ and with the Transformer, we obtain the raw scale and shift $[\boldsymbol{\alpha}^{(k)}, \boldsymbol{\beta}^{(k)}] \in \mathbb{R}^{H\times 2d_a}$. To stabilize training, we set the scale and shift to be
\begin{align}
\mathbf{b}^{(k)}_i &= 4\tanh\bigl(\frac{\boldsymbol{\beta}^{(k) }_i}{4}\bigr) \label{eq:tanhclip}\\
\mathbf{s}^{(k)}_i &= \log\left[1 + (e - 1)\exp{\left({4\tanh(\frac{\boldsymbol{\alpha}^{(k)}_i}{4})}\right)}\right]
\label{eq:softplus}
\end{align}
then the flow block yields
\begin{equation}
\begin{aligned}
\mathbf{v}^{(k)}_i
&=\bigl(\mathbf{u}^{(k)}_i- \mathbf{b}^{(k)}_i\bigr)
\oslash \mathbf{s}^{(k)}_i\\
\mathbf{X}^{(k)}&=P_k^{-1}\mathbf{V}^{(k)},
\end{aligned}
\label{eq:affine_forward}
\end{equation}
where $\oslash$ denotes element-wise division and $i\in\{1,\ldots,H\}$ indexes tokens in the block's processing order. The flow sets $[\boldsymbol{\alpha}^{(k)}_1, \boldsymbol{\beta}^{(k)}_1] = [\boldsymbol{0}, \mathbf{0}]$, which yields $[\mathbf{b}^{(k)}_1, \mathbf{s}^{(k)}_1] = [\boldsymbol{0}, \mathbf{1}]$, meaning identity transformation on $\mathbf{u}^{(k)}_1$.

\subsection{Maximum-likelihood training}
\label{sec:training}
The shifted autoregressive dependence makes each block's Jacobian block lower-triangular in processing order, with diagonal blocks $\operatorname{diag}(1/\mathbf{s}^{(k)}_i)$.
Since permutations have unit absolute determinant,
\begin{equation}
\log\left|\det J_{f_k}\right|
=-\sum_{i=1}^{H}\sum_{j=1}^{d_a}\log s^{(k)}_{i,j}.
\label{eq:block_logdet}
\end{equation}
We fit a Gaussian-smoothed distribution over augmented normalized action chunks:
\begin{equation}
\widetilde{\mathbf{X}}_t=\mathbf{X}_t+\sigma\boldsymbol{\epsilon},
\qquad
\boldsymbol{\epsilon}\sim\mathcal{N}(\mathbf{0},\mathbf{I}),
\label{eq:noise}
\end{equation}
where $\sigma$ controls the smoothing strength. Let $q_{\theta}(\widetilde{\mathbf{X}}_t\mid\mathbf{C}_t)$ denote the learned density of the perturbed actions.
With $\mathbf{X}^{(0)}=\widetilde{\mathbf{X}}_t$, $\mathbf{Z}=\mathbf{X}^{(K)}$, and demonstration dataset $\mathcal{D}$, we minimize the loss per scalar action dimension:
\begin{equation}
\mathcal{L}=\frac{1}{H d_a}\mathbb{E}_{\mathcal{D},\boldsymbol{\epsilon}}
\left[
\frac{1}{2}\|\mathbf{Z}\|_F^2
+\sum_{k=1}^{K}\sum_{i=1}^{H}\sum_{j=1}^{d_a}\log s^{(k)}_{i,j}
\right],
\label{eq:training_loss}
\end{equation}
which is the negative log-likelihood per scalar action dimension.
We jointly optimize the flow blocks, context projections, and vision--language backbone using this single objective, without auxiliary prediction or representation-alignment losses.

\subsection{Action generation and closed-loop execution}
\label{sec:inference}

At inference, we compute $\mathbf{C}_t$, sample a standard Gaussian latent chunk, and invert the flow:
\begin{equation}
\operatorname{vec}(\mathbf{Z})\sim\mathcal{N}(\mathbf{0},\mathbf{I}_{H d_a}),
\qquad
\widetilde{\mathbf{X}}_t=f_{\theta}^{-1}(\mathbf{Z};\mathbf{C}_t).
\label{eq:sampling}
\end{equation}
We invert the blocks in reverse order, $k=K,\ldots,1$, first applying the corresponding permutation $\mathbf{V}^{(k)}=P_k\mathbf{X}^{(k)}$.
The first processed token is unchanged, $\mathbf{u}^{(k)}_1=\mathbf{v}^{(k)}_1$, and subsequent tokens are recovered sequentially as
\begin{equation}
\mathbf{u}^{(k)}_i
=\mathbf{v}^{(k)}_i\odot\mathbf{s}^{(k)}_i
+\mathbf{b}^{(k)}_i,\qquad i=2,\ldots,H,
\label{eq:affine_inverse}
\end{equation}
where $\odot$ denotes element-wise multiplication and the coefficients are evaluated on the recovered prefix $\mathbf{U}^{(k)}_{<i}$.
In the deep block, we first prefix the attention key--value cache with the context tokens.
Feeding each recovered token $\mathbf{u}^{(k)}_{i-1}$, with earlier tokens cached, produces the coefficients for token $i$, matching the forward output shift.
Each block therefore requires $H-1$ sequential token evaluations.
We restore the original order with $\mathbf{X}^{(k-1)}=P_k^{-1}\mathbf{U}^{(k)}$ before inverting the next block; no classifier-free guidance is used.

The final output normalized action chunk is processed with one score-based denoising step to refine the sampled actions to account for the Gaussian smoothing used during training~\cite{efron2011tweedie,zhai2025tarflow}:
\begin{equation}
\widehat{\mathbf{X}}_t=
\widetilde{\mathbf{X}}_t+\sigma^2
\nabla_{\widetilde{\mathbf{X}}_t}
\log q_{\theta}(\widetilde{\mathbf{X}}_t\mid\mathbf{C}_t).
\label{eq:tweedie}
\end{equation}
For the exact Gaussian-smoothed data density, Tweedie's identity gives the conditional posterior mean of the clean normalized action chunk.
Substituting the learned density yields an approximate denoising estimate, not an exact sample from the clean action distribution.
We compute the score through an additional likelihood evaluation and differentiation with respect to the sampled actions, holding the context fixed.
No separate score network is required.
Finally, we recover action units with the action statistics
\begin{equation}
\widehat{\mathbf{A}}_t=
\widehat{\mathbf{X}}_t\odot\mathbf{s}_a+\boldsymbol{\mu}_a.
\label{eq:unnormalization}
\end{equation}

\section{Experiments}
\label{sec:experiments}

We evaluated whether exact maximum-likelihood training supports competitive manipulation performance on LIBERO and how this performance changes under the perturbations in LIBERO-Plus.
Both evaluations used the same NF-VLA checkpoint without additional fine-tuning on perturbed tasks.

\subsection{Implementation details}
\label{sec:implementation}

\paragraph{Observations and actions.}
We initialized the vision--language backbone from SmolVLM-500M-Instruct~\cite{marafioti2025smolvlm} and fine-tuned it jointly with the action model. Each input image was resized to $384\times384$ and encoded into 36 visual tokens.
The 72 visual tokens and 50 instruction-token positions produced 122 features of width 960.
Instructions were right-padded, with padded positions retained in the backbone computation.
After projection and prepending the proprioception token, the flow context had shape $123\times1024$.
We used action chunks with $H=10$ and $d_a=7$, comprising three translational increments, three rotational increments, and one gripper command per step. Actions and proprioception were normalized over each dimension using training-set statistics.

\paragraph{Flow configuration.}
The action head contained six affine autoregressive blocks: five shallow blocks with two Transformer layers each and one conditioned deep block with 14 layers.
All blocks used width 1024 and 16 attention heads.
One-dimensional rotary position embeddings~\cite{su2021roformer} used a frequency base of $10{,}000$.
In the deep block, context and action tokens occupied positions 0--122 and 123--132, respectively; shallow blocks used positions 0--9 in their processing order. Gaussian smoothing used $\sigma=0.05$ in normalized action space.

\paragraph{Training and inference.}
We trained the policy with the full LIBERO dataset on \textbf{4$\times$B200} GPUs with a global batch size of 256, starting from the pretrained VLM without robot pretraining.
At inference, we sampled at temperature 1.0, applied one Tweedie correction step and then unnormalized the actions.
Flow inversion cached attention keys and values and required $K(H-1)=54$ sequential token evaluations, in addition to context-prefix processing and the subsequent score computation.

\subsection{Evaluation setup}
\label{sec:evaluation_setup}

\paragraph{LIBERO.}
We evaluated the Spatial, Object, Goal, and Long suites of LIBERO~\cite{liu2023libero}, each containing ten tasks.
For each task, we ran 50 episodes, using each provided initial state once, for 500 episodes per suite and 2{,}000 episodes overall. 

\paragraph{LIBERO-Plus.}
We tested the policy's zero-shot generalization on 10{,}030 perturbed task instances spanning camera viewpoints, robot initial states, language instructions, lighting, background textures, sensor noise, and object layouts~\cite{fei2025liberoplus}.

\subsection{Performance on LIBERO}
\label{sec:libero_results}

\begin{table}[t]
\centering
\caption{\textbf{Comparison on the LIBERO benchmark.} We report success rate (\%) on LIBERO over all four task suits.
Success rate (\%, higher is better) on LIBERO. Avg. is the mean over the four listed suites. NF-VLA is evaluated for 50 trials per task. \textbf{Bold*} indicates the highest success rate in each suite, and \textbf{bold} indicates the second-highest.}
\label{tab:libero_results}
\begin{tabular}{lccccc}
\toprule
\textbf{Model} & \textbf{Spatial} & \textbf{Object} & \textbf{Goal} & \textbf{Long} & \textbf{Avg.} \\
\midrule
UniVLA~\cite{bu2025univla} & 96.5 & 96.8 & 95.6 & 92.0 & 95.2 \\
OpenVLA~\cite{kim2024openvla} & 84.7 & 88.4 & 79.2 & 53.7 & 76.5 \\
OpenVLA-OFT~\cite{kim2025oft} & \textbf{97.6} & 98.4 & 97.9 & \textbf{94.5} & \textbf{97.1}\\
$\pi_{0.5}$~\cite{black2025pi05} & \textbf{98.8*} & 98.2 & \textbf{98.0} & 92.4 & 96.9 \\
MolmoAct~\cite{lee2025molmoact} & 87.0 & 95.4 & 87.6 & 77.2 & 86.6 \\
CoT-VLA~\cite{zhao2025cotvla} & 87.5 & 91.6 & 87.6 & 69.0 & 81.1 \\
$\pi_0$~\cite{black2024pi0} & 96.8 & 98.8 & 95.8 & 85.2 & 94.2 \\
WorldVLA~\cite{cen2025worldvla} & 87.6 & 96.2 & 83.4 & 60.0 & 81.8 \\
UnifiedVLA~\cite{wang2025unifiedvla} & 95.4 & 98.8 & 93.6 & 94.0 & 95.5 \\
SmolVLA~\cite{shukor2025smolvla} & 93.0 & 94.0 & 91.0 & 77.0 & 88.8 \\
UniAct~\cite{zheng2025uniact} & 77.0 & 87.0 & 77.0 & 70.0 & 77.8 \\
NinA (Transformer)~\cite{tarasov2025nina} & 97.0 & 97.8 & 93.8 & 92.0 & 95.2 \\
RoMAN-Flow (NF-IQL) ~\cite{romanflow} & 95.0 &\textbf{99.4} & 94.6 & 92.2 & 95.3 \\
\midrule
NF-VLA (ours) & 97.4 & \textbf{99.8*} & \textbf{98.4*} & \textbf{97.8*} & \textbf{98.4*} \\
\bottomrule
\end{tabular}
\end{table}

NF-VLA achieved 98.35\% success on LIBERO, completing 1{,}967 of 2{,}000 episodes (Table~\ref{tab:libero_results}).
Among the listed models, NF-VLA achieved the highest overall score, 1.5 percentage points above $\pi_{0.5}$'s reported 96.85\%.
It also achieved the highest scores on Object, Goal, and Long, including 97.8\% on Long compared with $\pi_{0.5}$'s 92.4\%, while scoring 1.4 points lower on Spatial.
These results support the viability of the likelihood-trained action model across the evaluated tasks, including longer action sequences. Of the 33 failed episodes, 11 occurred in two drawer-related tasks.

\subsection{Robustness on LIBERO-Plus}
\label{sec:libero_plus_results}

\begin{table}[t]
\centering
\caption{\textbf{Comparison on the LIBERO-Plus benchmark.} We report success rate (\%) on LIBERO-Plus for each and overall average across all perturbations. \textbf{Bold*} indicates the highest success rate in each column, and \textbf{bold} indicates the second-highest.Baseline results are taken from LIBERO-Plus~\cite{fei2025liberoplus}.}
\label{tab:libero_plus_results}
\resizebox{\linewidth}{!}{%
\begin{tabular}{lcccccccc}
\toprule
\textbf{Model} & \textbf{Camera} & \textbf{Robot} & \textbf{Language} & \textbf{Light} & \textbf{Background} & \textbf{Noise} & \textbf{Layout} & \textbf{Avg.} \\
\midrule
OpenVLA~\cite{kim2024openvla} & 0.8 & 3.5 & 23.0 & 8.1 & 34.8 & 15.2 & 28.5 & 15.6 \\
OpenVLA-OFT~\cite{kim2025oft} & 56.4 & 31.9 & \textbf{79.5} & 88.7 & \textbf{93.3} & 75.8 & \textbf{74.2} & \textbf{69.6} \\
OpenVLA-OFT$_{\mathrm{w}}$~\cite{kim2025oft,fei2025liberoplus} & 10.4 & 38.7 & 70.5 & 76.8 & \textbf{93.6*} & 49.9 & 69.9 & 55.8 \\
NORA~\cite{hung2025nora} & 2.2 & 37.0 & 65.1 & 45.7 & 58.6 & 12.8 & 62.1 & 39.0 \\
WorldVLA~\cite{cen2025worldvla} & 0.1 & 27.9 & 41.6 & 43.7 & 17.1 & 10.9 & 38.0 & 25.0 \\
UniVLA~\cite{bu2025univla} & 1.8 & \textbf{46.2} & 69.6 & 69.0 & 81.0 & 21.2 & 31.9 & 43.9 \\
$\pi_0$~\cite{black2024pi0} & 13.8 & 6.0 & 58.8 & 85.0 & 81.4 & \textbf{79.0*} & 68.9 & 53.6 \\
$\pi_0$-Fast~\cite{pertsch2025fast} & \textbf{65.1*} & 21.6 & 61.0 & 73.2 & 73.2 & 74.4 & 68.8 & 61.6 \\
RIPT-VLA~\cite{tan2025riptvla} & 55.2 & 31.2 & 77.6 & 88.4 & 91.6 & 73.5 & \textbf{74.2} & 68.4 \\
OpenVLA-OFT$_{\mathrm{m}}$~\cite{kim2025oft,fei2025liberoplus} & 55.6 & 21.7 & \textbf{81.0*} & \textbf{92.7} & 91.0 & \textbf{78.6} & 68.7 & 67.9 \\
\midrule
NF-VLA (ours) & \textbf{57.0} & \textbf{53.8*} & 64.8 & \textbf{96.9*} & 87.4 & 75.8 & \textbf{83.7*} & \textbf{72.6*} \\
\bottomrule
\end{tabular}%
}
\end{table}
We also evaluated NF-VLA's zero-shot generalization on the LIBERO-Plus benchmark without any fine-tuning (Table~\ref{tab:libero_plus_results}).
NF-VLA achieved 72.6\% overall success, with 53.8\% on robot initial states, 96.9\% on lighting, and 83.7\% on object layouts.

\section{Ablations}
\label{sec:ablations}

\begin{figure}
\centering
\includegraphics[width=\linewidth]{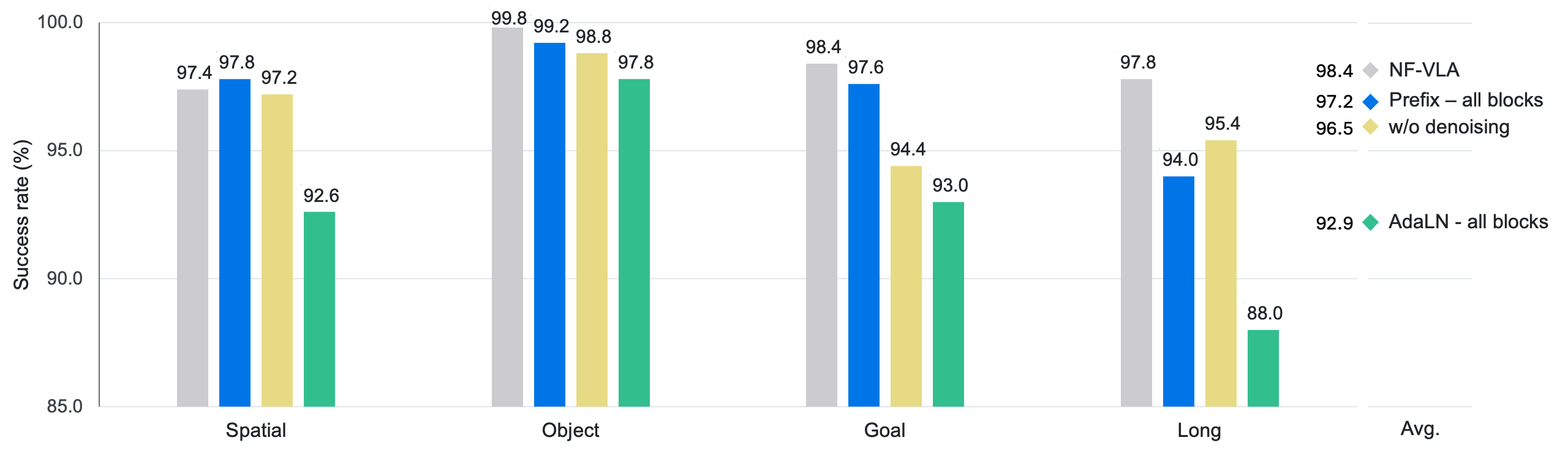}
\caption{\textbf{Effect of NF-VLA design choices on LIBERO success.} Bars show success rates for the baseline and three ablation variants across the four suites; diamonds show suite averages rounded to one decimal place. Prefix and AdaLN variants condition both the deep and shallow blocks.}
\label{fig:ablations}
\end{figure}

We examined the contributions of score-based denoising, conditioning depth, and conditioning mechanism (Figure~\ref{fig:ablations}).
The full NF-VLA configuration from Table~\ref{tab:libero_results} used VLM features as prefix context in the deep block, left the shallow blocks unconditioned, and applied score-based denoising after generation.
The conditioning variants retained this denoising step.

\paragraph{Score-based denoising.}
Removing the denoising step from the trained NF-VLA checkpoint reduced average success by 1.90 percentage points.
The largest decrease occurred on Goal, where success fell from 98.4\% to 94.4\%.
This suggests the score-based denoising rectifies the action generation from the learned Gaussian-smoothed density and improves downstream control performance.

\paragraph{Conditioning depth.}
Extending VLM prefix conditioning to the shallow blocks and retraining reduced average success by 1.20 percentage points relative to the full model.
Spatial success increased slightly, but performance declined on the other three suites, with the largest decrease on Long, from 97.8\% to 94.0\%.
Thus, conditioning the deep block alone achieved higher overall success than distributing prefix context across all flow blocks in this evaluation.
This result is consistent with the design rationale of concentrating direct conditioning in the high-capacity deep block (Section~\ref{sec:flow_architecture}).

\paragraph{Conditioning mechanism.}
Replacing prefix conditioning with adaptive layer normalization (AdaLN)~\cite{Peebles_2023_ICCV} in both deep and shallow blocks yielded 92.85\% average success.
Compared with prefix conditioning in all blocks, AdaLN reduced average success by 4.30 percentage points and performed worse on every suite.
This comparison holds conditioning depth fixed and supports prefix conditioning over AdaLN for the evaluated action model.

\paragraph{Inference efficiency.}
We benchmarked the three conditioning configurations in FP32 with BF16 autocast on a single \textbf{RTX 4080 SUPER} GPU, measuring NF sampling and the Tweedie step separately (Table~\ref{tab:inference_efficiency}).
Relative to AdaLN, prefix conditioning reduced NF sampling latency by 30.3\% for the baseline and 25.9\% when applied to all blocks.
Both prefix variants were faster  in our implementation despite attending to more context tokens, as AdaLN modulation introduced additional computation.
Including VLM inference (12.11$\pm0.01$) ms, NF-VLA achieves an end-to-end inference latency of 74.10 ms per action chunk.

\begin{table}[htbp]
\centering
\caption{\textbf{Inference efficiency of conditioning variants.} Latencies were measured in FP32 with BF16 autocast.}
\label{tab:inference_efficiency}
\small
\setlength{\tabcolsep}{5pt}
\begin{tabular}{lcccc}
\toprule
 & \shortstack{\textbf{NF sampling (ms)}} & \shortstack{\textbf{Tweedie step (ms)}} & \shortstack{\textbf{GPU mem. (GB)}} & \shortstack{\textbf{NF param. (M)}} \\
\midrule
NF-VLA (baseline) & $50.10 \pm 0.03$ & $11.89 \pm 0.02$ & 2.75 & 303 \\
Prefix - all blocks & $53.29 \pm 0.07$ & $13.16 \pm 0.01$ & 2.79 & 303 \\
AdaLN - all blocks & $71.90 \pm 0.24$ & $12.95 \pm 0.01$ & 3.36 & 483 \\
\bottomrule
\end{tabular}
\end{table}

\section{Conclusion}
\label{sec:conclusion}

We studied conditioning and sampling in autoregressive-flow VLA policies through NF-VLA, a baseline trained end to end with a single exact maximum-likelihood objective on Gaussian-smoothed actions.
The learned density supports both action sampling and score-based correction.
In the evaluated configurations, restricting prefix conditioning to the deep flow block achieved higher average success than conditioning all blocks, and prefix conditioning outperformed AdaLN at the same all-block placement.
A single Tweedie correction improved closed-loop success using the model's own score, without additional training or changes to the objective.
Together with the reported inference costs, these results establish the practical value of conditioning placement and self-score correction within this policy family.
NF-VLA achieved strong task performance with a perturbation-dependent robustness profile.
These findings support autoregressive NFs as a basis for studying how multimodal context and learned action densities can be used for robot control.

\bibliography{nfvla_references}

\end{document}